\documentclass[12pt, letterpaper]{article}
\usepackage[margin=2cm]{geometry}
\usepackage{changes}

\usepackage{amssymb}
\usepackage{amsmath}
\usepackage{graphicx}

\usepackage{mathtools}
\usepackage{url}

\usepackage{cite}

\usepackage{rotating}

\usepackage{nicefrac}   

\usepackage{color}

\usepackage[ruled,vlined]{algorithm2e}

\usepackage[utf8]{inputenc}

\usepackage{titlesec}
\titleclass{\subsubsubsection}{straight}[\subsubsection]
\newcounter{subsubsubsection}[subsubsection]
\renewcommand\thesubsubsubsection{\thesubsubsection.\arabic{subsubsubsection}}
\titleformat{\subsubsubsection}
  {\normalfont\normalsize\bfseries}{\thesubsubsubsection}{1em}{}
\titlespacing*{\subsubsubsection}
  {0pt}{3.25ex plus 1ex minus .2ex}{1.5ex plus .2ex}

\title{Mamba time series forecasting with uncertainty quantification}
\author{Pedro Pessoa$^{1,2}$,  Paul Campitelli$^{1,2}$, Douglas P. Shepherd$^{1,2}$, \\ S. Banu Ozkan$^{1,2}$, Steve Press\'e$^{1,2,3}$ \\
$^1$Center for Biological Physics, 
$^2$Department of Physics,\\
$^3$  School of Molecular Sciences, \\ 
Arizona State University, 
Tempe, AZ, USA
}
\date{}

\begin{document}

\maketitle

\abstract{
State space models, such as Mamba, have recently garnered attention in time series forecasting due to their ability to capture sequence patterns. 
However, in electricity consumption benchmarks, Mamba forecasts exhibit a mean error of approximately 8\%. Similarly, in traffic occupancy benchmarks, the mean error reaches 18\%. This discrepancy leaves us to wonder whether the prediction is simply inaccurate or falls within error given spread in historical data. To address this limitation, we propose a method to quantify the predictive uncertainty of Mamba forecasts.
Here, we propose a dual-network framework based on the Mamba architecture for probabilistic forecasting, where one network generates point forecasts while the other estimates predictive uncertainty by modeling variance. We abbreviate our tool, Mamba with probabilistic time series forecasting, as Mamba-ProbTSF and the code for its implementation is available on GitHub \url{https://github.com/PessoaP/Mamba-ProbTSF}.
Evaluating this approach on synthetic and real-world benchmark datasets, we find Kullback-Leibler divergence between the learned distributions and the data--which, in the limit of infinite data, should converge to zero if the model correctly captures the underlying probability distribution--reduced to the order of $10^{-3}$ for synthetic data and $10^{-1}$ for real-world benchmark, demonstrating its effectiveness. 
We find that in both the electricity consumption and traffic occupancy benchmark, the true trajectory stays within the predicted uncertainty interval at the two-sigma level about 95\% of the time.   
We end with a consideration of potential limitations, adjustments to improve performance, and considerations for applying this framework to processes for purely or largely stochastic dynamics where the stochastic changes accumulate, as observed for example in pure Brownian motion or molecular dynamics trajectories.
}

%\newpage
\section{Introduction}

Time series forecasting (TSF) is the task of predicting how to complete sequences. In other words, inferring--or forecasting--additional elements of a sequence given a subset of its preceding elements\cite{Sezer20,Raksha21,DoblasReyes13,Rama22,Benitez21,Wang24,Ma24}. At its core, TSF operates under the assumption that the observed sequence is a realization of some underlying dynamical process\cite{Presse10,Bryan20,Bryan22,Kilic23,Kilic21,Pessoa24,Pessoa21}.  This perspective suggests that if we can learn the governing laws driving these dynamics from the available portion of the sequence, the past, we can extend our understanding to forecast the future.

Mathematically, we define the lookback horizon of ``past" values as follows: $\{x_1, x_2, \ldots, x_P\}$, and abbreviate to $x_{1:P}$. Our goal is to predict the next $T$ (the forecast horizon) values, denoted $x_{P+1:P+T}$. This is achieved through a function represented by a $T$-variate forecast function ${f_{1:T}}(x_{1:P}) = \left(f_1(x_{1:P}), \ldots, f_T(x_{1:P})\right)$, such that $x_{P+\tau} \approx f_\tau(x_{1:P})$. 
This can be achieved with an explicit dynamical model; for example, in the case of a free-falling particle, we can use past observations to calculate parameters such as initial velocity, gravitational acceleration, and relevant drag coefficients. 
However, in many fields where TSF is applied -- such as finance\cite{Sezer20} and climate science\cite{Raksha21,DoblasReyes13} -- we lack access to explicit dynamical models, or the governing equations may be too complex to describe analytically.

Under these scenarios, rather than assuming a predefined model, we attempt to learn the model. For this task, neural networks (NNs) provide a framework capable of capturing high-dimensional relationships within the data \cite{Goodfellow16}. These NNs are trained by being exposed to multiple realizations of the same underlying  dynamical process, observing sequences of past and future data points, and optimizing a convex loss function of the difference between the predicted and the actual future values. 

\begin{figure}
     \centering
     \includegraphics[width=0.9\linewidth]{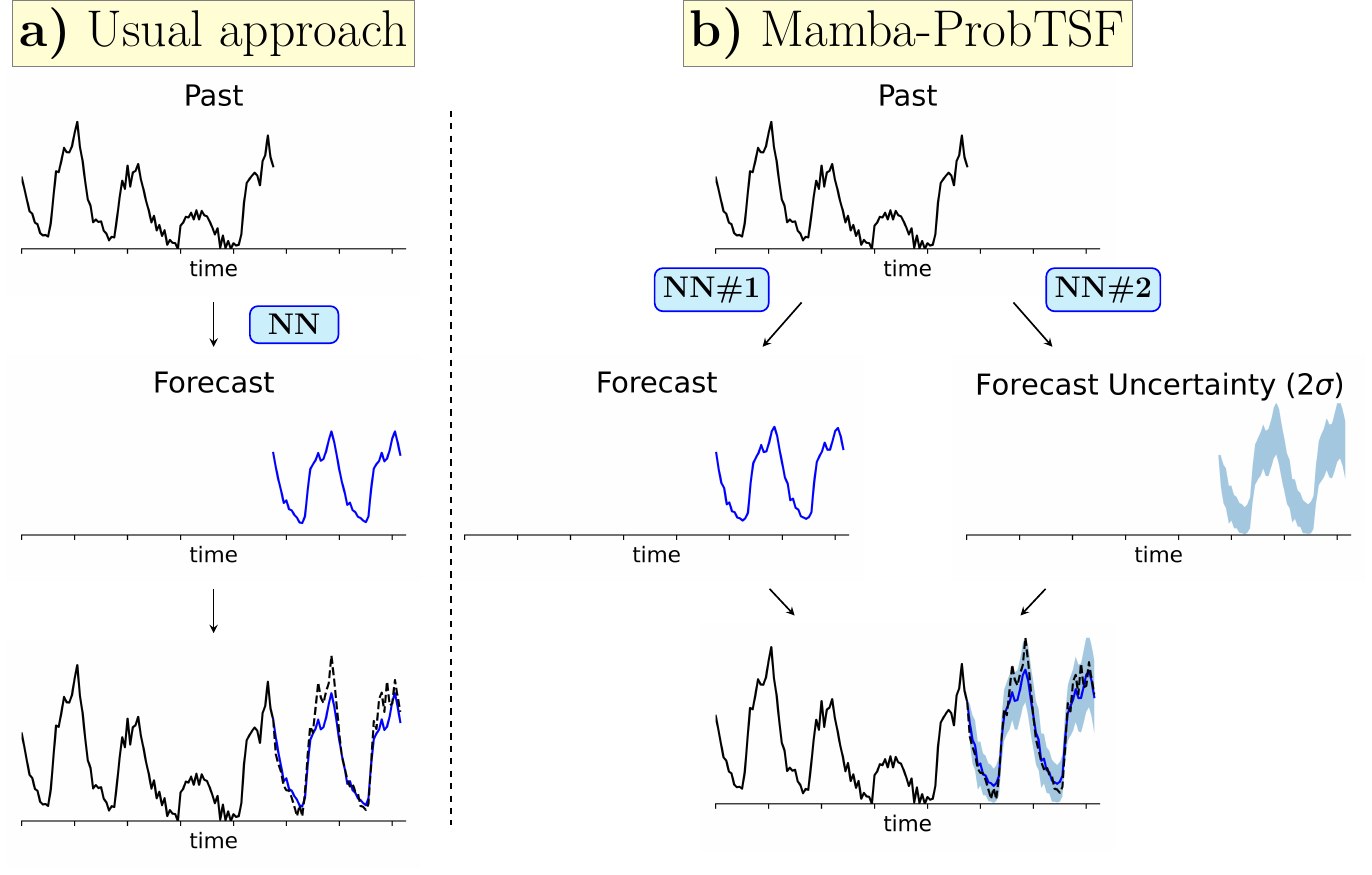}
     \caption{\textbf{Summary of our dual-network approach in Mamba-ProbTSF.} a) The standard deterministic approach and b) our proposed dual-network probabilistic approach. In the standard approach, a single neural network forecasts future values directly from historical data, typically minimizing point-wise errors.
     In our approach, two separate neural networks are employed: one forecasts the future trajectory, while the other estimates the associated uncertainty by leveraging correlations across the dataset at different time points observed during training.
     As shown in this example -- derived from the electricity dataset analysis detailed in Sec. \ref{sec:electricity_chapter} -- this structure enables smoother forecasts by appropriately attributing variability to uncertainty. As a result,  the real future trajectory (dashed black line) lies within the  uncertainty forecast.}\label{fig:graphical_abstract}
 \end{figure}

More precisely, the NN approximates the forecast function as  ${f_{1:T}}^\phi(x_{1:P})$ where $\phi$ represents the internal network parameters. These parameters are optimized to minimize a loss function, such as   $\sum\limits_{\tau=1}^T \left(x_{P+\tau} - f^\phi_\tau(x_{1:P}) \right)^2. $ 
This approach is summarized in Fig. \ref{fig:graphical_abstract}a.

The form of $ f^\phi $ is determined by the network architecture. While selecting the appropriate architecture remains a challenging task in the field of NNs, state space models (SSMs) -- and in particular, the Mamba architecture \cite{dao24,Gu24} -- have emerged as highly successful for TSF \cite{Wang24,Ma24,mambastock}. Mamba combines the efficiency of SSMs with an advanced selective mechanism akin to attention \cite{Attention}  outperforming transformer-based architectures in both accuracy and computational efficiency \cite{Wang24}.
Though these models provide point predictions for future values, they do not inherently quantify the uncertainty associated with these predictions. 
Under many practical scenarios, knowing how much one can trust a prediction is just as important as the prediction itself \cite{Bryan20,Kilic21,Bryan22,Saurabh23,Rojewski23,Takavoli20,Sgouralis24,Pessoa24}.
These arguments, in turn, motivate our probabilistic extension of Mamba.

Existing NN approaches that yield non-deterministic forecasts are often generative models,
where repeated evaluations of the same input produce distinct future trajectories \cite{AutoBNN,Pearce18}.
While this variability captures predictive uncertainty, it is unclear whether the differences between trajectories represents uncertainty in the underlying dynamics \cite{LeGuen20}. Thus, there is no guarantee that the spread of trajectories aligns with the uncertainty in the dynamics under study informed from historical data.
Therefore, rather than predicting multiple possible trajectory continuations, the objective should be to learn a probability distribution over the ``future'' conditioned on the observed ``past'' $p(x_{P+1:P+T}| x_{1:P})$. In other words, develop a probabilistic TSF\cite{LeGuen20,GluonTS,Saad23,Wang23}.

In this work, we adopt a more structured approach to TSF in a probabilistic manner by introducing a dual-network architecture designed to estimate both the forecasted trajectory and its forecasted uncertainty explicitly thus making it possible to calculate probability densities of the forecast given the past. The implementation, termed Mamba-ProbTSF, is available in our GitHub repository\cite{github}. 
Specifically, we define two functions
\begin{equation}\label{NNfunctions}
\begin{aligned}
{\mu_{1:T}}^{\phi_\mu}(x_{1:P}) &= \left(\mu^{\phi_\mu}_1(x_{1:P}), \ldots, \mu^{\phi_\mu}_T(x_{1:P})\right), \quad  \text{and} \\
{\sigma_{1:T}}^{\phi_\sigma}(x_{1:P}) &= \left(\sigma^{\phi_\sigma}_1(x_{1:P}), \ldots, \sigma^{\phi_\sigma}_T(x_{1:P})\right),
\end{aligned}
\end{equation} 
where $\mu^{\phi_\mu}_\tau$ represents the predicted (or forecast) value at step $\tau$, and $\sigma^{\phi_\sigma}_\tau$ represents the associated uncertainty (or forecast uncertainty) at step $\tau$. 
To enforce probabilistic consistency, we assume that the conditional distribution of future values is Gaussian centered at the predicted mean with variance given by the squared predicted standard deviation  -- that is $x_{P+\tau} | x_{1:P} \sim \text{Normal}\left(\mu^{\phi_\mu}_\tau (x_{1:P}), \left(\sigma^{\phi_\sigma}_\tau (x_{1:P}) \right)^2\right)$. 
A summary of this approach is given in Fig. \ref{fig:graphical_abstract}b. 
Any correlation between future points is determined by the past, as the future values $x_{P+\tau}$ are assumed conditionally independent given $x_{1:P}$. In other words, the covariance between different time steps $\tau$ is assumed to be diagonal.

In the following sections, we evaluate the proposed architecture using both synthetic and real-world datasets. 
This is done by analyzing the standardized residuals -- meaning the difference between the real future data $x_{P:P+T}$, and the forecast ${\mu_{1:T}}^{\phi_\mu}(x_{1:P})$, divided by the forecast uncertainty ${\sigma_{1:T}}^{\phi_\sigma}(x_{1:P})$, later defined in Sec. \ref{sec:metrics} -- across a testing dataset. 
Ideally, when the network is properly trained and the stochasticity in the underlying model is indeed Gaussian, as discussed in the examples of Sec. \ref{sec:Results}, these standardized residuals should follow a standard Gaussian distribution. To quantify the model's performance, we examine the variance of these standardized residuals and compute their Kullback-Leibler (KL) divergence from a standard normal distribution.
In what follows, we observe the success of this approach as well as its limitations.

%\newpage
\section{Methods}

\subsection{State-Space Models}\label{sec:ssm}
State-of-the-art neural network models for time series forecasting (TSF) are built upon state-space models (SSMs)\cite{Wang23,Ma24,Wang24}. While a comprehensive review of all SSM architectures in the literature is beyond the scope of this article, we describe the mathematical foundations of Mamba, a specific SSM, along with its implicit assumptions. This discussion provides insights into Mamba's applicability, which, in turn, are useful in interpreting the results presented in Sec. \ref{sec:Results}.

SSMs operate under the premise that the observed data is influenced by an underlying latent process, which the model attempts to infer. To describe SSMs in a general context, we index the input, $\mathcal{I}_t^n$, with $t$ denoting time and $n$ denoting different trajectories in the dataset each corresponding to a distinct realization of a same underlying dynamical process. 

The SSM produces the output $\mathcal{O}_t^n$ in two steps. The first step applies a differential equation model to the latent variable $h^n$:
\begin{equation}
    \frac{\mathrm{d}h^n(t)}{\mathrm{d}t} = \sum_{m=1}^N \left( A^n_m h^m(t) + B^n_m \mathcal{I}_t^m \right),
\end{equation}
where $N$ denotes the number of trajectories in the input and both $A^n_m$ and $B^n_m$ are learned matrices, meaning each element is a parameter of the neural network optimized during training. The sum over $m$ accounts for interactions across multiple latent variables, reflecting a coupled system where each trajectory $n$ is influenced not only by its own past but also by shared information from others. This allows the model to leverage dependencies across different trajectories particularly useful for time series exhibiting common underlying dynamics. The second step projects the latent state onto the output space
\begin{equation}\label{ssm-dynamic}
    \mathcal{O}_t^n = \sum_{m=1}^N C^n_m h^m(t),
\end{equation}
where $C^n_m$ is another learnable matrix responsible for the final projection. 

In practice, these continuous-time equations must be discretized for implementation in modern neural network architectures. The discretization step transforms \eqref{ssm-dynamic} into:
\begin{equation}
    h^n_{t+1} = \sum_{m=1}^N \left( \tilde{A}^n_m h^m_t + \tilde{B}^n_m \mathcal{I}_t^m\right),
\end{equation}
where $\tilde{A}^n_m$ and $\tilde{B}^n_m$ are discrete approximations of the continuous parameters, given in terms of $A^n_m$ and $B^n_m$ through the matrix exponential
\begin{equation}
    \tilde{A}^n_m = \left[\exp(\Delta_t\mathbf{A} )\right]^n_m, \quad \text{and} \quad  
    \tilde{B}^n_m = \sum_{o=1}^N \left[ (\Delta_t \mathbf{A})^{-1} (\exp(\Delta_t \mathbf{A}) - \mathbf{I}) \right]^n_o B^o_m.
\end{equation}
Here, $\mathbf{A}$ represents the matrix of elements $A^n_m$, $\mathbf{I}$ is the identity matrix, and $\Delta_t$ 
is a learnable real-valued discretization step.
This transformation ensures numerical stability and preserves the underlying system dynamics over discrete time intervals without assuming that the time between observations is small enough to warrant a first order approximation.

By discretizing the state-space formulation, SSMs can be efficiently implemented in neural architectures, leveraging parallelization techniques to process long sequences while maintaining the model's structural advantages. Integrating learned parameterizations with discretized updates allows models like Mamba to achieve both flexibility and scalability, making them particularly effective for time-series forecasting tasks that require capturing long-range dependencies. For further details, we refer to the original Mamba authors \cite{dao24,Gu24}. 
For now, we note that SSMs  expect a latent process described as a differential equation and, as a consequence, follow a deterministic dynamics.

\subsection{Network training}\label{sec:training}
Here, we describe how to train the two networks, representing ${\mu_{1:T}}^{\phi_\mu}(x_{1:P})$ and ${\sigma_{1:T}}^{\phi_\sigma}(x_{1:P})$ in \eqref{NNfunctions}, for probabilistic TSF. 
For the standard architecture for ${\mu_{1:T}}^{\phi_\mu}(x_{1:P})$, we use S-Mamba \cite{Wang24}, while for ${\sigma_{1:T}}^{\phi_\sigma}(x_{1:P})$, we use a fully connected neural network with a softplus activation in the last layer to ensure a strictly positive output. 
The full implementation is available in our GitHub repository \cite{github}.

\subsubsection{Loss function for probabilistic TSF}
As mentioned in the introduction, we assume a Gaussian conditional probability for a point in the ``future" conditioned on the ``past" given by  
\begin{equation}\label{prob_perpoint}
        p(x_{P+\tau}|x_{1:P})  = \frac{1}{\sqrt{2\pi} \ \sigma^{\phi_\sigma}_\tau(x_{1:P})} \exp\left[ -  \frac{1}{2} \left( \frac{x_{P+\tau} - \mu^{\phi_\mu}_\tau(x_{1:P})}{ \sigma^{\phi_\sigma}_\tau(x_{1:P})} \right)^2 \right]  ,
\end{equation}
and the neural network training consists of finding the parameter sets ${\mu_{1:T}}^{\phi_\mu}$ and ${\sigma_{1:T}}^{\phi_\sigma}$ that best represent the underlying dynamics. While the assumption of a Gaussian probability distribution at every point aligns well with many scientifically relevant scenarios -- \emph{e.g.}, Brownian motion (further discussed in Sec. \ref{sec:brownian}) \cite{Sgouralis24} or deterministic dynamics with Gaussian observational noise (further discussed in Sec. \ref{sec:sines} and \ref{sec:vdp}) \cite{Bryan20,Bryan22}. Maximum entropy also provides a justification: the Gaussian distribution is the least biased choice when only the first and second moments of the distribution are specified \cite{Jaynes03,Presse13,Caticha12,Pessoa21}. Thus, even in cases where the Gaussian assumption may not strictly hold, the Gaussian-inspired models allows the dual-network to still learn the predictions mean (forecast) and variance (forecast uncertainty). 

Specifically, we partition some dynamical realizations into a training dataset, where each trajectory is indexed by a superscript $n \in \{1, \ldots , N\}$, such that each trajectory in the dataset is given as $x^n_{1:P+T}$.  
To estimate the optimal parameters, we maximize the total conditional likelihood, assuming independence between trajectories
\begin{equation}\label{joint-independence}
        p(x_{P+1:P+T}^{1:N}|x_{1:P}^{1:N})  = \prod_{n=1}^N  p(x_{P+1:P+T}^n|x_{1:P}^n)
        =        \prod_{n=1}^N  \prod_{\tau=1}^T p(x_{P+\tau}^n|x_{1:P}^n)  .
\end{equation}
Although the values of $x$ at different $\tau$ are treated as conditionally independent, meaning there is no explicit correlation imposed between them in the probability model, the functions $\mu^{\phi_\mu}_\tau(x_{1:P})$ and $\sigma^{\phi_\sigma}_\tau(x_{1:P})$ depend on the full past trajectory $x_{1:P}$. As a result, these functions can implicitly capture temporal correlations across times within the forecast horizon, allowing the model to learn dependencies in the data despite the per-step independence assumption of \eqref{joint-independence}.  
This is equivalent to minimizing the negative log-likelihood, leading to the following loss function: 
\begin{equation}\label{gen-loss}
  \mathcal{L}(\phi_\mu,\phi_\sigma) = \frac{1}{N} \sum_{n=1}^N \  \sum_{\tau=1}^{T} \left[ \frac{1}{2} \left( \frac{x_{P+\tau}^n - \mu^{\phi_\mu}_\tau(x_{1:P}^n)}{ \sigma^{\phi_\sigma}_\tau(x_{1:P}^n)} \right)^2 + \log \sigma^{\phi_\sigma}_\tau(x_{1:P}^n)\right] ,
\end{equation}
which applies the full probabilistic forecasting.

\subsubsection{Steps of training}\label{sec:training_steps}
With the joint loss function \eqref{gen-loss} specified, a few more details of the training process are in order. 
First, when we initialize the networks, we train the network for $\mu^{\phi_\mu}_{1:T}(x_{1:P})$ separately from $\sigma^{\phi_\sigma}_{1:T}(x_{1:P})$.

We pre-initialize the first network parameters, $\phi_\mu$, by training them as if we had a non-probabilistic TSF procedure. That is, we train the first network by using only a point-wise error as the loss function, meaning we minimize
\begin{equation}\label{loss_pre}
    \mathcal{L}_{pre}(\phi_\mu) = \frac{1}{N} \sum_{n=1}^N \sum_{\tau=1}^{T} \left( x_{P+\tau}^n - \mu^{\phi_\mu}_\tau(x_{1:P}^n) \right)^2.
\end{equation}
This initialization ensures that the mean predictor ${\mu_{1:T}}^{\phi_\mu}$ learns a stable estimate of the expected future trajectory before incorporating uncertainty. This is then equivalent to  \eqref{gen-loss} if one assumed that the uncertainty at all times is equal, that is $\sigma^{\phi_\sigma}_\tau(x_{1:P}^n) = C$, for all $\tau$ and $n$.

Once the initialization of $\phi_\mu$ is complete, we proceed with the full probabilistic training. Since training with all trajectories simultaneously may exceed memory limitations, we employ minibatching. That is, in each training step, we randomly select a batch of $B$ trajectories from the dataset and compute the loss:
\begin{equation}
  \mathcal{L}_B(\phi_\mu,\phi_\sigma) = \frac{1}{B} \sum_{n \in \mathcal{B}} \sum_{\tau=1}^{T} \left[ \frac{1}{2} \left( \frac{x_{P+\tau}^n - \mu^{\phi_\mu}_\tau(x_{1:P}^n)}{ \sigma^{\phi_\sigma}_\tau(x_{1:P}^n)} \right)^2 + \log \sigma^{\phi_\sigma}_\tau(x_{1:P}^n) \right].
\end{equation}
where $\mathcal{B}$ represents the minibatch of size $B$. After computing the minibatch loss, we update parameters $\phi_\mu$ and $\phi_\sigma$ using gradient-based optimization tools within PyTorch. 

\subsection{Metrics of success}\label{sec:metrics}

A common challenge in probabilistic TSF is to verify whether the learned distribution accurately represents the intrinsic stochastic dynamics under study. However, our data is provided in such a way that, for each given ``past" we observe only a single corresponding ``future." In other words, we have only one sample from each conditional probability distribution.  
As such, some approaches attempt to train networks using synthetic datasets, where multiple future samples are generated for a single past \cite{LeGuen20}. However, this strategy is not consistent with how data is typically presented in real TSF tasks, where only one realization of the stochastic dynamics is available for each historical sequence.  

Here, we adopt a different method to quantify whether the neural network has learned the correct distribution. Fortunately, since our training is based on a Gaussian conditional probability \eqref{prob_perpoint}, we can define the standardized residual at step $\tau$:  
\begin{equation}\label{Z-def}
    z_\tau^n =  \frac{x_{P+\tau}^n - \mu^{\phi_\mu}_\tau(x_{1:P}^n)}{ \sigma^{\phi_\sigma}_\tau(x_{1:P}^n)}.
\end{equation}  
This quantity represents the normalized deviation of the observed value from the predicted mean, scaled by the predicted standard deviation at step $\tau$.  

If the Gaussian assumption holds for the dynamics under study and the dual network accurately learn the dynamics, the standardized residuals $z_\tau^n$ should follow a standard Normal distribution, with mean 0 and variance 1, for all $\tau$.  
This property provides a straightforward way to assess the model's performance: we check whether the empirical distribution of $z_\tau^n$ across all $\tau$ matches the expected standard normal distribution. Here, $n$ indexes the trajectories in the testing set, as this evaluation is performed on the test data, not during training. Unlike the previous subsection, where $n$ referred to the training data, here we use $n$ to denote unseen test samples.

From the standardized residuals $z_\tau^n$, we define two key metrics of success. The first is the variance of $z_\tau^n$, which should ideally approach 1 for each $\tau$. The second is the Kullback-Leibler (KL) divergence\cite{Goodfellow16} between the empirical distribution of $z_\tau^n$ and the standard normal distribution, which should ideally approach zero. These metrics provide both a quantitative and qualitative assessment of model calibration.

In the following section, when describing each dataset under study, we will present histograms of $z_\tau^n$ at different values of $\tau$: specifically, at $\tau = 1$, at the maximum $\tau = T$, and at the midpoint $\tau = \lfloor T/2 \rfloor$. Qualitatively, these histograms should resemble the shape of a standard normal distribution. Additionally, we will show how both the variance and KL divergence evolve as a function of $\tau$, providing insights into how well the model maintains its probabilistic consistency across different forecast horizons.

\section{Results}\label{sec:Results}

In this section, we discuss the results of the probabilistic time series forecasting (TSF) approach proposed in the previous section, applied to both real and synthetic datasets. Each dataset will be described in its respective subsection.

\subsection{Sines (synthetic)}  \label{sec:sines}
As a first example, we generate a simple synthetic dataset where each data point is constructed as a sum of sine waves with different frequencies, plus independent Gaussian noise. Each trajectory in the dataset is given as
\begin{equation}\label{sines_data}
    x_t^n = 4 \sin(\omega_1 t+ \phi^n) + \sin(\omega_2^n t ) + \xi^n_t.
\end{equation}
We set $\omega_1 = 2\pi/24$, while for each trajectory, $n$, we have a single $\omega_2^n$ is sampled from an exponential distribution with a mean $ 2\pi/12$ and initial phase $\phi^n$ sampled uniformly between 0 and $2\pi$. The noise term $\xi^n_t$ is independently drawn from a standard normal distribution for all $t$. 
To clarify notation, $t$ indexes all data points across both the ``past" and ``future" $(1:P+T)$, whereas in the previous section, $\tau$ was used to refer specifically to indices within the ``future" range, with $\tau$ running from $0$ to $T$.

\begin{figure}
    \centering
    \includegraphics[width=0.95\linewidth]{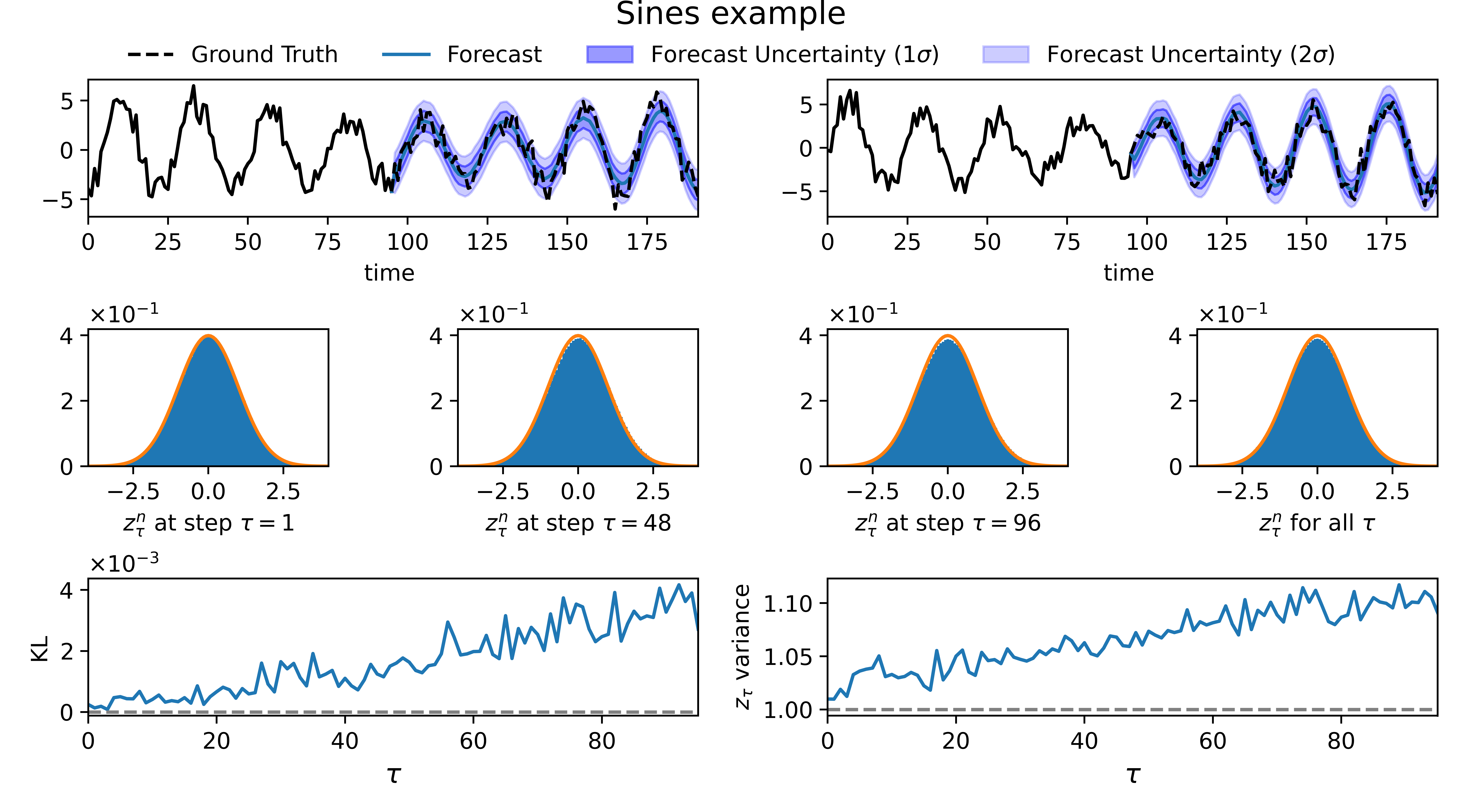}\vspace{-.5cm}
    \caption{\textbf{Application of Mamba-ProbTSF  to a sum of sines with added Gaussian noise}.  
    The first row shows a sample trajectory along with its forecast and predicted uncertainty.  
    The middle row presents histograms of standardized residuals $z_\tau^n$ at the beginning of the forecast ($\tau=1$), the midpoint, the final forecast point ($\tau = 96$), and the histogram where we pool all values of $z_\tau^n$ across all $\tau$,} compared to a standard normal distribution (orange line).  
    The bottom row demonstrates that the variance of $z_\tau^n$ remains close to 1 across different forecast horizons $\tau$.  
    The KL divergence between the empirical and standard normal distributions remains low, on the order of $10^{-3}$. 
    \label{fig:sines}
\end{figure}

The results obtained from this dataset are shown in Fig. \ref{fig:sines}. We observe that the dual-network system successfully captures both underlying dynamics, as illustrated by the histograms of $z^n_\tau$ matching standard Gaussians, KL approaching 0 (with values of the order of $10^{-3}$) and variance near 1. 

\subsection{Van der Pol Dynamics (synthetic)}\label{sec:vdp}
As a second synthetic example, we generate a dataset based on the van der Pol oscillator \cite{VdP20}, which exhibits nonlinear deterministic dynamics and is a prototype model for systems with self-excited limit cycle oscillations and has been applied to various physical and biological phenomena\cite{Lee13,Vaid15,Inaba24}. 

Each trajectory in the dataset is obtained by numerically integrating the van der Pol differential equation for an underlying variable $y(t)$ 
\begin{equation}\label{vdp_ode}
    \frac{d^2 y}{dt^2} - \omega_1^2 \left(\lambda^n (1 - y^2) \frac{dy}{dt} + y \right) = 0.
\end{equation}
We set $\omega_1 = 2\pi/24$, while for each trajectory $n$, the damping coefficient $\lambda^n$ is sampled from an exponential distribution with mean $5$. The system is numerically integrated using a fixed time step, with initial conditions $y_0 = 0$ and $dy/dt \vert_{t=0} = 1$. The observed dataset is then constructed by sampling the integrated solution at integer time steps and adding independent Gaussian noise
\begin{equation}
    x_t^n = y_t^n + \xi_t^n,
\end{equation}
where $\xi_t^n$ is sampled independently from a standard normal distribution for all $t$.  

This setup mirrors the approach used in the sines example, where the observed data combines a deterministic underlying process with stochastic noise. 
Unlike the sines example, which follows simple harmonic motion, the van der Pol system exhibits nonlinear oscillations where the amplitude and frequency depend on the damping parameter $\lambda^n$. This makes it a valuable benchmark in testing forecasting models on more complex, state-dependent dynamics.

\begin{figure}
    \centering
    \includegraphics[width=0.95\linewidth]{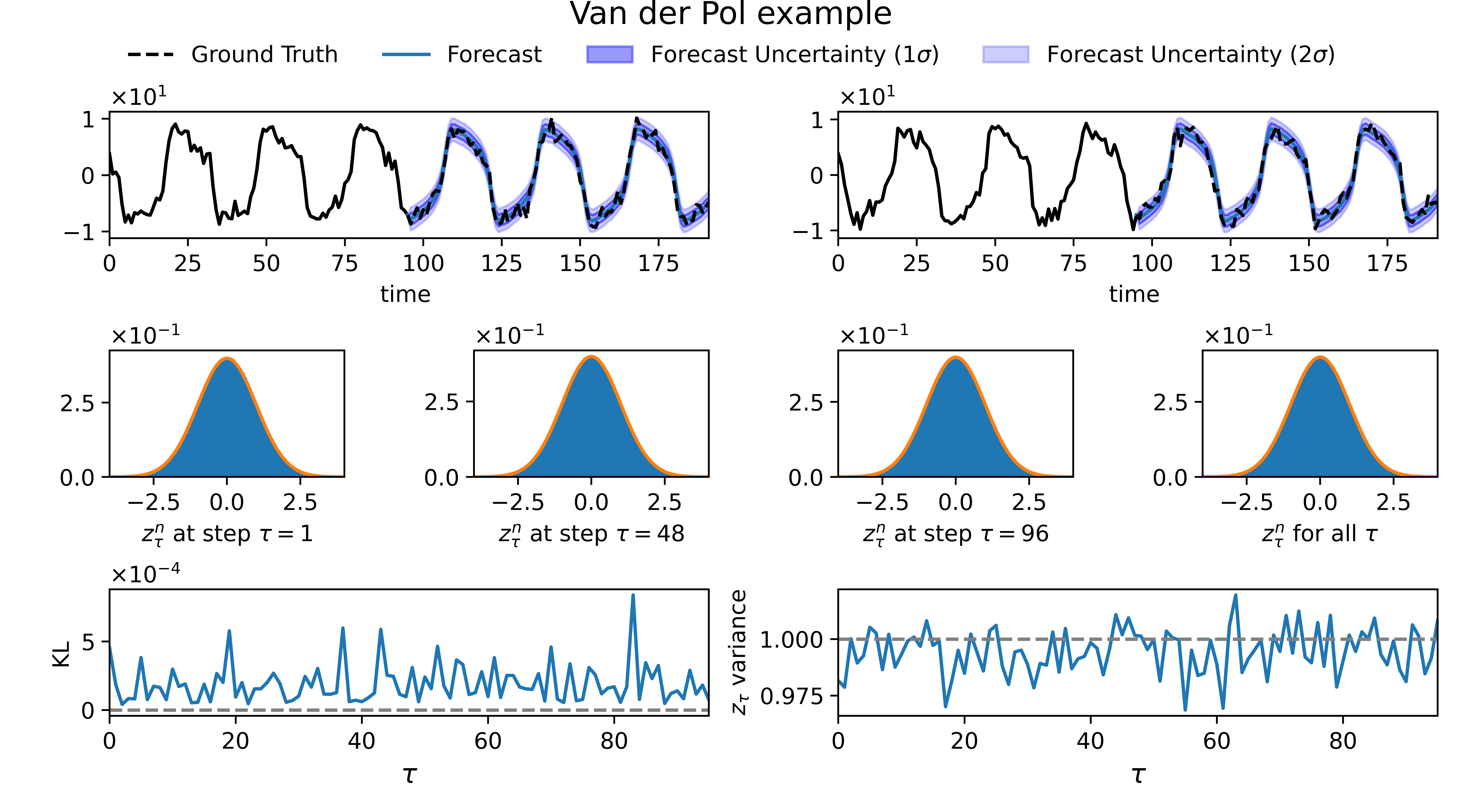}
    \vspace{-.5cm}
    \caption{\textbf{Application of {Mamba-ProbTSF} to the van der Pol oscillator with added Gaussian noise}.  
    Similarly to Fig. \ref{fig:sines}, the first row shows a sample trajectory with its forecast and predicted uncertainty.  
    The middle row presents histograms of standardized residuals $z_\tau^n$ at $\tau=1$, the midpoint, $\tau = 96$, and all values of $\tau$ pooled together compared to a standard normal distribution (orange line).  
    The bottom row shows that the variance of $z_\tau^n$ remains close to 1 across different forecast horizons $\tau$.  
    The KL divergence between the empirical and standard normal distributions is even lower than in the sines example, on the order of $10^{-4}$.}
    \label{fig:VdP}
\end{figure}

The results obtained from this dataset are shown in Fig. \ref{fig:VdP}. Compared to the sines example, we observe that the dual-network system successfully captures the more complex underlying dynamics. This is reflected in the histograms of $z^n_\tau$, where the KL divergence approaches 0 (on the order of $10^{-4}$), and the variance remains close to 1, similar to the sines example.

\subsection{Electricity consumption (real-world)}\label{sec:electricity_chapter}

As a real-world example, we consider the Electricity dataset\cite{Trindade15}, a widely used benchmark in TSF \cite{Wang24, Challu23, Wu22}. This dataset records the hourly electricity consumption of 321 customers from 2012 to 2014, originally collected at fifteen-minute intervals \cite{Challu23} and later aggregated for consistency in analysis.

Unlike synthetic datasets, which follow predefined equations, electricity consumption arises from a complex and partially observable dynamical system. The underlying dynamics are influenced by multiple interacting factors, including consumer behavior, economic activity, weather conditions, and energy policies, all of which evolve over time. Despite this complexity, the system exhibits periodicity at daily and weekly scales due to human routines, alongside irregular fluctuations driven by external influences.

\begin{figure}
    \centering
    \includegraphics[width=0.95\linewidth]{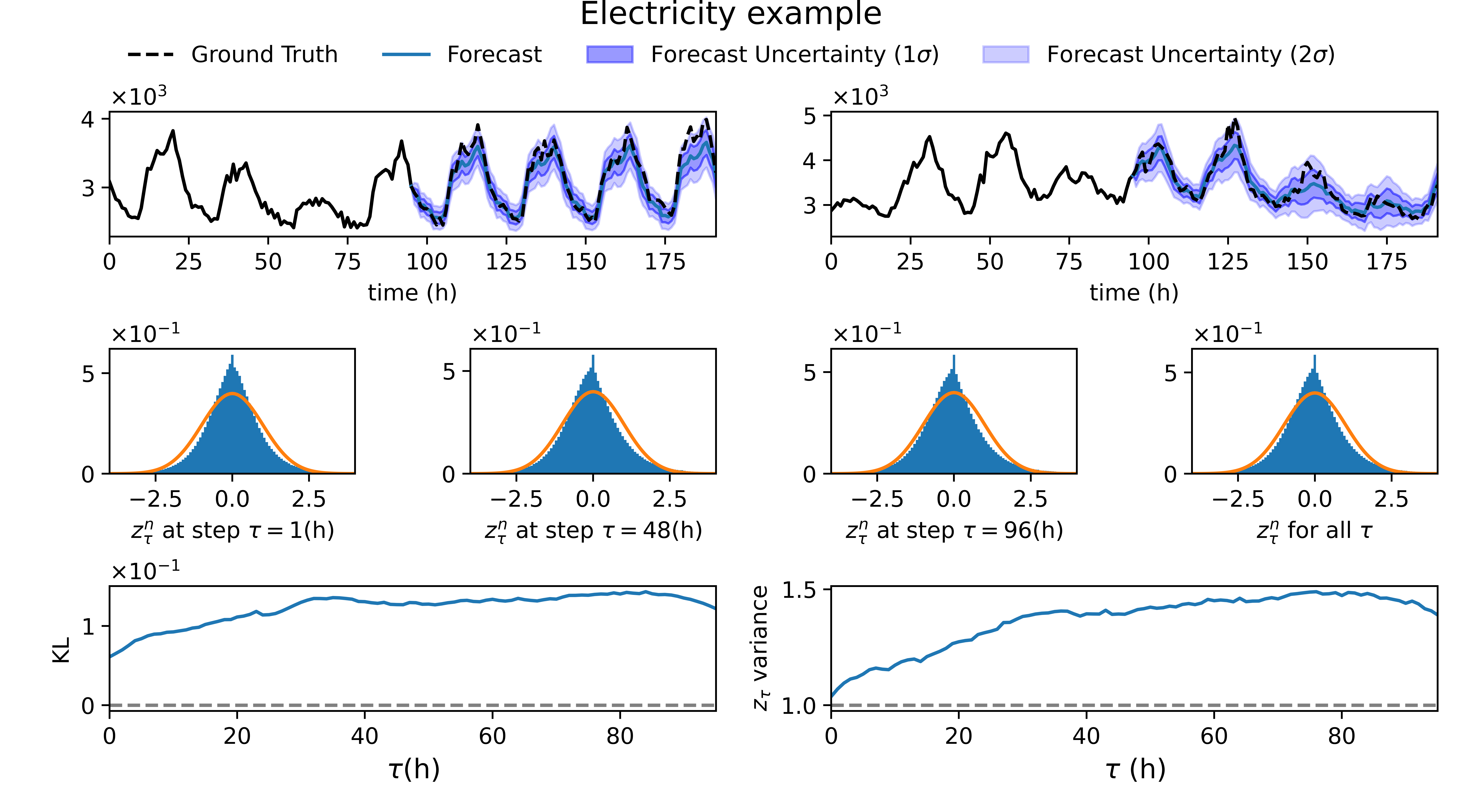}
    \vspace{-.5cm}
    \caption{\textbf{Mamba-ProbTSF applied to the electricity real-world dataset highlights the need for uncertainty modeling}.  The figure presents results using a fully connected neural network for $\sigma^{\phi_\sigma}_\tau(x_{1:P})$, ensuring stable uncertainty estimates.  The probabilistic TSF approach achieves a close match between the standardized residuals, $z_\tau^n$, and the expected standard normal distribution, with a KL divergence on the order of $10^{-1}$.  The variance remains within a reasonable range, varying with $\tau$ but staying below 1.5, demonstrating a well-learned and stable uncertainty model.
    }
    \label{fig:electricity}
\end{figure}

In Fig. \ref{fig:electricity}, we present the results of the dual-network strategy for the electricity dataset. We observe that the KL divergence between standardized residuals $z_\tau^n$ and a standard normal distribution remains stable, on the order of $10^{-1}$, indicating that the learned uncertainty quantification aligns well with the empirical data. Additionally, the variance of $z_\tau^n$ across different forecast horizons remains below $1.5$, demonstrating a consistent and reasonable estimate of forecast uncertainty.

\begin{figure}
    \centering
    \includegraphics[width=0.95\linewidth]{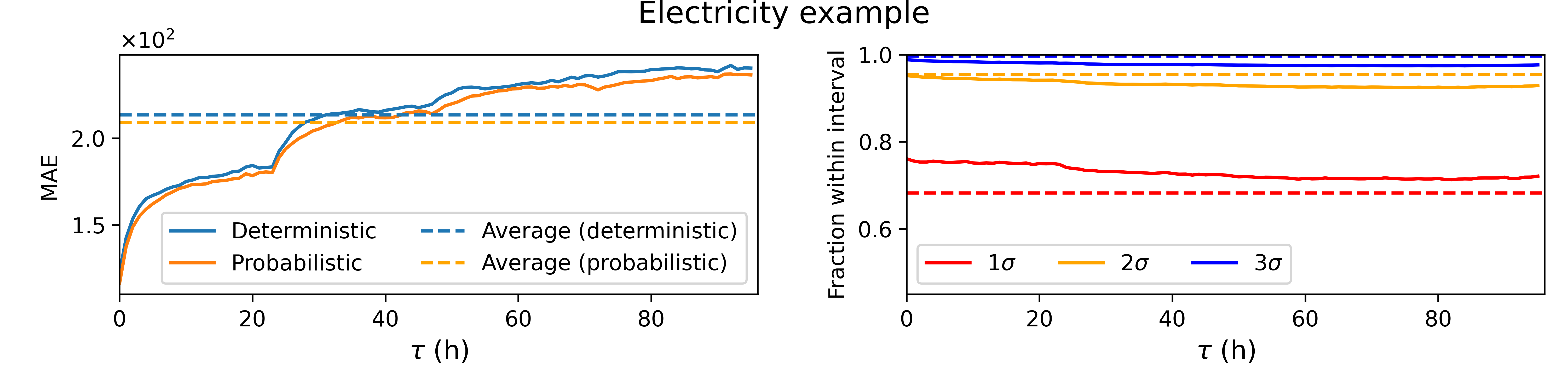}
    \vspace{-.5cm}
\caption{\textbf{Probabilistic forecasting provides strong error quantification.}  
(Left) Mean absolute error (MAE) for deterministic forecasts averages 213 Wh per residence, while the probabilistic approach trivially reduces this to 209 Wh per residence, compared to a total average of 2.65 kWh in hourly measurements (or, equivalently, 63.6 kWh in a day).  
(Right) Empirical coverage of uncertainty intervals: solid lines depict the fraction of test trajectories within the $1\sigma$, $2\sigma$, and $3\sigma$ uncertainty bounds, while dashed lines indicate the expected coverage under a perfect Gaussian assumption (68.3\%, 95.4\%, and 99.7\%, respectively). The actual empirical coverage over time averages to 73\% for the $1\sigma$ interval, 93\% for the $2\sigma$ interval, and 98\% for the $3\sigma$ interval.
These results indicate that, for most of the dataset, the model provides a conservative uncertainty estimate, particularly for smaller forecast horizons $\tau$.}  
    \label{fig:electricity2}
\end{figure}

To evaluate the reliability of uncertainty estimates, we measure the fraction of trajectories in the test dataset that fall within the forecast uncertainty bounds (Fig. \ref{fig:electricity2}).
Under an ideal Gaussian assumption, 68.3\% of data points should fall within the $1\sigma$ forecast uncertainty interval ($ \mu^{\phi_\mu}_\tau(x_{1:P}^n) \pm \sigma^{\phi_\sigma}_\tau(x_{1:P}^n)$), while 95.4\% should fall within the $2\sigma$ interval.
However, in practice, real trajectories stay within the $1\sigma$ uncertainty interval 72.9\% of the time on average and within the $2\sigma$ interval 93\% of the time. This suggests that when applied to real-world data, where the Gaussian assumption in \eqref{prob_perpoint} is not strictly valid, the Gaussian-based loss function tends to provide a conservative estimation of uncertainty.
Therefore, the model’s uncertainty bounds slightly overestimate risk, which may be beneficial for cautious decision-making, such as using Mamba forecasts to manage energy grid responses.

To assess the impact of probabilistic forecasting in this real-world benchmark, we compare the mean absolute error (MAE) of deterministic S-Mamba forecasts to those of our probabilistic approach (Fig. \ref{fig:electricity2}).  
Interestingly, the probabilistic model achieves a slight improvement in MAE, reducing it from 213 Wh to 209 Wh per residence.  
This improvement arises from the model’s ability to adapt uncertainty estimates across different trajectories and time steps.  
While deterministic training, based on \eqref{loss_pre}, implicitly assumes a constant error magnitude across all instances, the probabilistic training approach, which optimizes \eqref{gen-loss}, relaxes this assumption by allowing the uncertainty distribution to dynamically adjust across different trajectories and time steps.

For completeness, it is important to mention that using S-Mamba for $\sigma^{\phi_\sigma}_{1:T}(x_{1:P})$ leads to overestimated uncertainty, indicating that the dynamics of uncertainty are not captured by a SSM. These results are presented in Supplemental Information \ref{SIsec:mamba_sigma}.

\subsection{Traffic Occupancy (Real-World)}\label{sec:traffic_chapter}

As another real-world benchmark, we consider the Traffic dataset \cite{Cal_traffic}, which consists of hourly measurements collected by the California department of transportation. This dataset records road occupancy rates from multiple sensors deployed along San Francisco Bay Area freeways and has been widely used in TSF research \cite{Wang24, Wu22}.

Similar to electricity consumption, traffic occupancy reflects human-driven dynamics but is subject to even greater external variability. While periodic patterns emerge from daily commuting routines, unpredictable disruptions such as city events, accidents, and weather conditions play an even more significant role in shaping traffic flow. 

\begin{figure}
    \centering
    \includegraphics[width=0.95\linewidth]{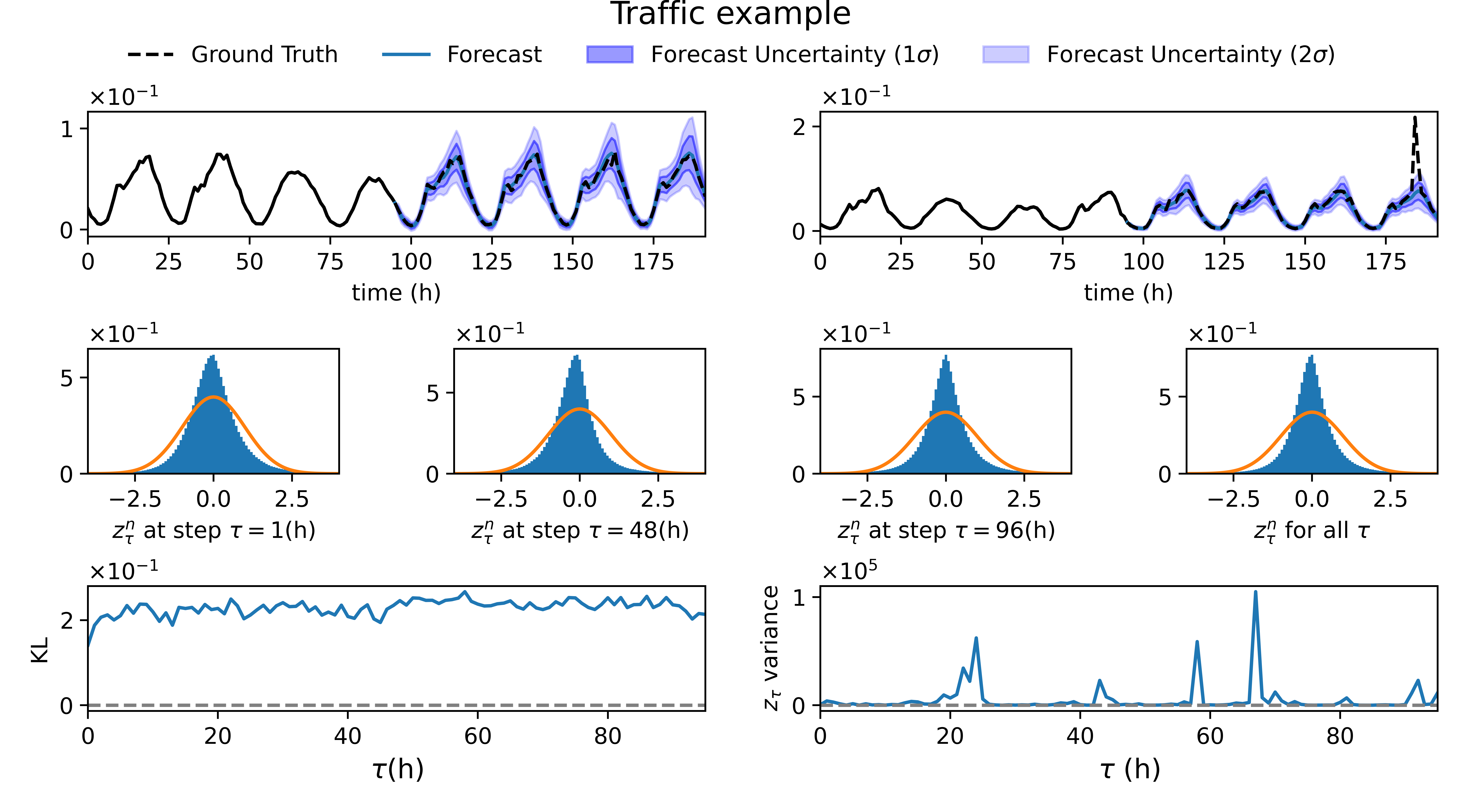}
    \vspace{-.5cm}
    \caption{\textbf{Mamba-ProbTSF in real-world traffic data.}
    The dual-network strategy ensures stable uncertainty estimates using a fully connected neural network for $\sigma^{\phi_\sigma}_\tau(x_{1:P})$.  
    The KL divergence between standardized residuals, $z_\tau^n$, and a standard normal distribution remains on the order of $10^{-1}$.  
    However, some rare congestion events cause high variance, as seen in the trajectory example (top-right), where a sudden traffic spike occurs.  
    Despite these anomalies, the model maintains reasonable and conservative uncertainty estimates.}
    \label{fig:traffic}
% \end{figure}  
% \begin{figure}
%     \centering
    \vspace{.5cm}
    \includegraphics[width=0.95\linewidth]{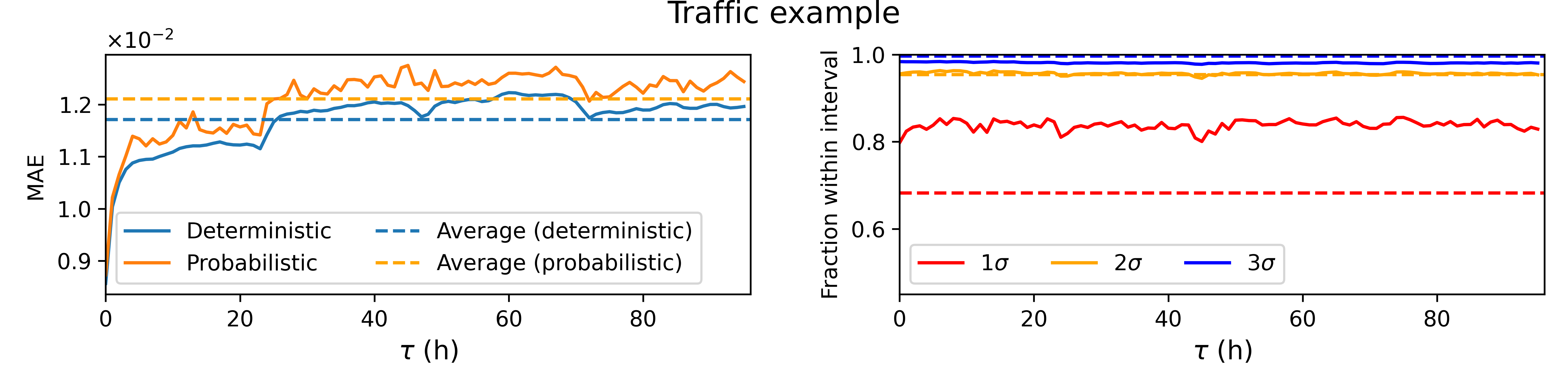}
    \vspace{-.5cm}
    \caption{\textbf{Empirical uncertainty coverage in traffic forecasting.}  
    (Left) Mean absolute error (MAE) increases slightly from 0.011 to 0.012, compared to an average total occupancy of 0.058, resulting in an 18\% error rate.  
    (Right) The solid lines show empirical uncertainty coverage, with 84\% of real trajectories falling within the $1\sigma$ interval and 96\% within the $2\sigma$ interval.  
    Compared to the electricity dataset, the model produces even more conservative uncertainty estimates.}
    \label{fig:traffic2}
\end{figure}

In Fig. \ref{fig:traffic}, we present results from the dual-network strategy applied to the Traffic dataset. The KL divergence between standardized residuals, $z_\tau^n$, and a standard normal distribution remains on the order of $10^{-1}$. However, the variance of $z_\tau^n$ can become significantly large, which can be attributed to rare but extreme congestion events. This is evident in the trajectory example (top-right panel of Fig. \ref{fig:traffic}), where a sudden traffic spike occurs due to such an event.

Despite the presence of these outliers, the model maintains conservative uncertainty estimates.  
In Fig. \ref{fig:traffic2}, we find that, on average, 84\% of real trajectories fall within the $1\sigma$ uncertainty interval, while 96\% remain within the $2\sigma$ interval. As such, when compared to the electricity dataset, the dual-network strategy produces even more conservative estimates, reflecting the higher volatility in traffic dynamics.  
Interestingly, in the top-left panel of Fig. \ref{fig:traffic}, we observe a case where periodicity is present, but the traffic magnitude decreases during the lookback horizon before increasing again. The model successfully captures this behavior, correctly predicting the eventual rise in traffic levels.

\subsection{Failure mode: Brownian motion (synthetic)}
\label{sec:brownian}

In this synthetic example, we consider the Brownian motion which presents a failure mode for SSMs. Brownian motion is a stochastic process where randomness is an inherent part of the system’s evolution rather than an external source of noise added on top of deterministic dynamics. Each trajectory in the dataset follows the recursive update:
\begin{equation}
    x_t^n = x_{t-1}^n + \xi_t^n,
\end{equation}
where each initial condition, $x_0^n$, is sampled from a uniform distribution on $[0,1]$, and each $\xi_t^n$ is sampled independently from a standard normal distribution for all $t$.  

Unlike the previous synthetic datasets, where the underlying system was deterministic with stochastic perturbations, Brownian motion is entirely driven by stochasticity. This introduces a key theoretical feature: given a past trajectory $x_{1:P}$, the conditional probability density function of future values follows:
\begin{equation}\label{brownian_real}
    p(x_{P+\tau} | x_{1:P}) = \frac{1}{\sqrt{2\pi \tau}} \exp\left[ -\frac{1}{2} \frac{\left(x_{P+\tau} - x_P\right)^2}{\tau}  \right].
\end{equation}
This result implies that the expected value remains at the last observed point, $x_P$, while the variance grows proportionally to $\tau$. 
This explicitly fits the Gaussian assumption as when we compare  with \eqref{prob_perpoint}, we identify  $\mu^{\phi_\mu}_\tau(x_{1:P}) \approx x_P$ for all $\tau$ and $\sigma^{\phi_\sigma}_\tau(x_{1:P}) \approx \sqrt{\tau}$.

The results obtained from this dataset are shown in the Supplemental Information  \ref{SIsec:brownian}. 
We evaluate the model's ability to capture the expected distribution using both S-Mamba and a fully connected linear network for $\sigma^{\phi_\sigma}_{1:T}$. In both cases, we find that the model struggles to fully recover the analytical distribution, with deviations particularly pronounced at both short and long forecasting horizons.

The explanation for this behavior is that SSMs such as Mamba are designed primarily for capturing deterministic latent structures and may not be well-suited for representing stochasticity as an intrinsic part of the dynamic. This highlights the challenge of learning purely stochastic processes with state-space models, the dynamical assumptions described in Sec. \ref{sec:ssm} are insufficient to fully account for the accumulation of uncertainty over time in processes such as Brownian motion. 
In other words, a dynamics like the one presented in \eqref{brownian_real}, will not be represented as differential equation and linear projection alone as expected by state space models.

\section{Discussion}

Here, we propose a dual-network architecture for TSF generating both the forecast and its associated uncertainty, brought together in Mamba-ProbTSF \cite{github}. The model is trained using a loss function based on the negative log-likelihood of a Gaussian distribution \eqref{gen-loss} where one network predicts the mean and the other estimates the square root of the variance. This formulation provides a forecast uncertainty that aligns with a Gaussian-like distribution.  

We evaluate the performance of this on two synthetic datasets characterized by known dynamics where the observations are given with additive Gaussian noise, as presented in Figs. \ref{fig:sines} and \ref{fig:VdP}. In those cases, we can observe a near perfect (measured by KL divergence and residue variance) match with the underlying distribution. 
In the real-world examples (Figs. \ref{fig:electricity} and \ref{fig:traffic}) we also obtain a good match (KL divergence on the order of $10^{-1}$) when using S-Mamba for the forecast network and a fully connected network for the uncertainty. 
While the match was considerably worse than in the synthetic cases, this can be attributed to the fact that real-world dataset variations are not strictly Gaussian. Nevertheless, by allowing the uncertainty estimates to adapt across different time points we can now capture 93\% and 96\% of the dataset within the $2\sigma$ interval in the eletricity (Fig. \ref{fig:electricity2}) and traffic (Fig. \ref{fig:traffic2}) benchmarks, respectively.

On the other hand, when applied to a process where the dynamics accumulate stochastic variations, such as Brownian motion. This suggests that while the proposed architecture effectively models uncertainty in systems with structured dynamics, additional considerations may be needed to extend its applicability to processes where  latent deterministic dynamics do not underlie the process itself. 
When dynamics are dominated by the accumulation of stochastic noise, it cannot be writen as the solution of a hidden differential equation, a fundamental assumptions of SSMs discussed in Sec. \ref{sec:ssm}. Addressing this challenge would require rethinking SSM formulations to better accommodate Brownian-like behavior, where uncertainty evolves as a function of stochastic accumulation rather than latent deterministic dynamics. Moreover, this enables the study of mixed cases in which the system follows deterministic dynamics while being influenced by Brownian-like noise \cite{Bryan20,Leimkuhler15,Lelivre16}, as seen in many approaches to molecular dynamics \cite{Zhu22,Janson23}.

Yet, given the strong results on structured datasets and real-world data, our study demonstrates the potential of the dual-network approach for probabilistic TSF. Future work could explore hybrid architectures, where the SSM is complemented by another modeling approach that better captures stochastic accumulation. In that sense NN approaches designed to approximate probability distributions, such as normalizing flows \cite{Rezende15,Durkan19,Stimper23,Pessoa25}, could further enhance the model's robustness in highly stochastic regimes.

\section*{Acknowledgments}
We would like to thank Zihan Wang and Sijie Huang for their assistance in setting up the computational environment and Julian Antolin Camarena for interesting discussions. 
SP acknowledges support from the NIH (R35GM148237), ARO (W911NF-23-1-0304), and NSF (Grant No. 2310610).

\bibliographystyle{elsarticle-num} 
\bibliography{refs}

\newpage
 \appendix
 \section*{Supplemental Information}
\renewcommand{\thepage}{S  -- \arabic{page}}
\setcounter{page}{1}
\pagenumbering{arabic} 

\renewcommand{\thefigure}{S\arabic{figure}}
\setcounter{figure}{0}
\renewcommand{\thetable}{S\arabic{table}}
\setcounter{table}{0}

 \section{Results using S-Mamba for variance}\label{SIsec:mamba_sigma}

 This Supplemental Information section we analyze the results of using the S-Mamba architecture for the uncertainty, mentioned as a possibility in Sec. \ref{sec:training} and as a possible failure for the real-world datasets. We replicate the results from Figs.  \ref{fig:sines},\ref{fig:VdP}, \ref{fig:electricity}, and \ref{fig:traffic} now using S-Mamba for $\sigma^{\phi_\sigma}_{1:T}(x_{1:P})$ instead of a fully connected neural network.

\begin{sidewaysfigure}
    \centering
    \includegraphics[width=0.45\linewidth]{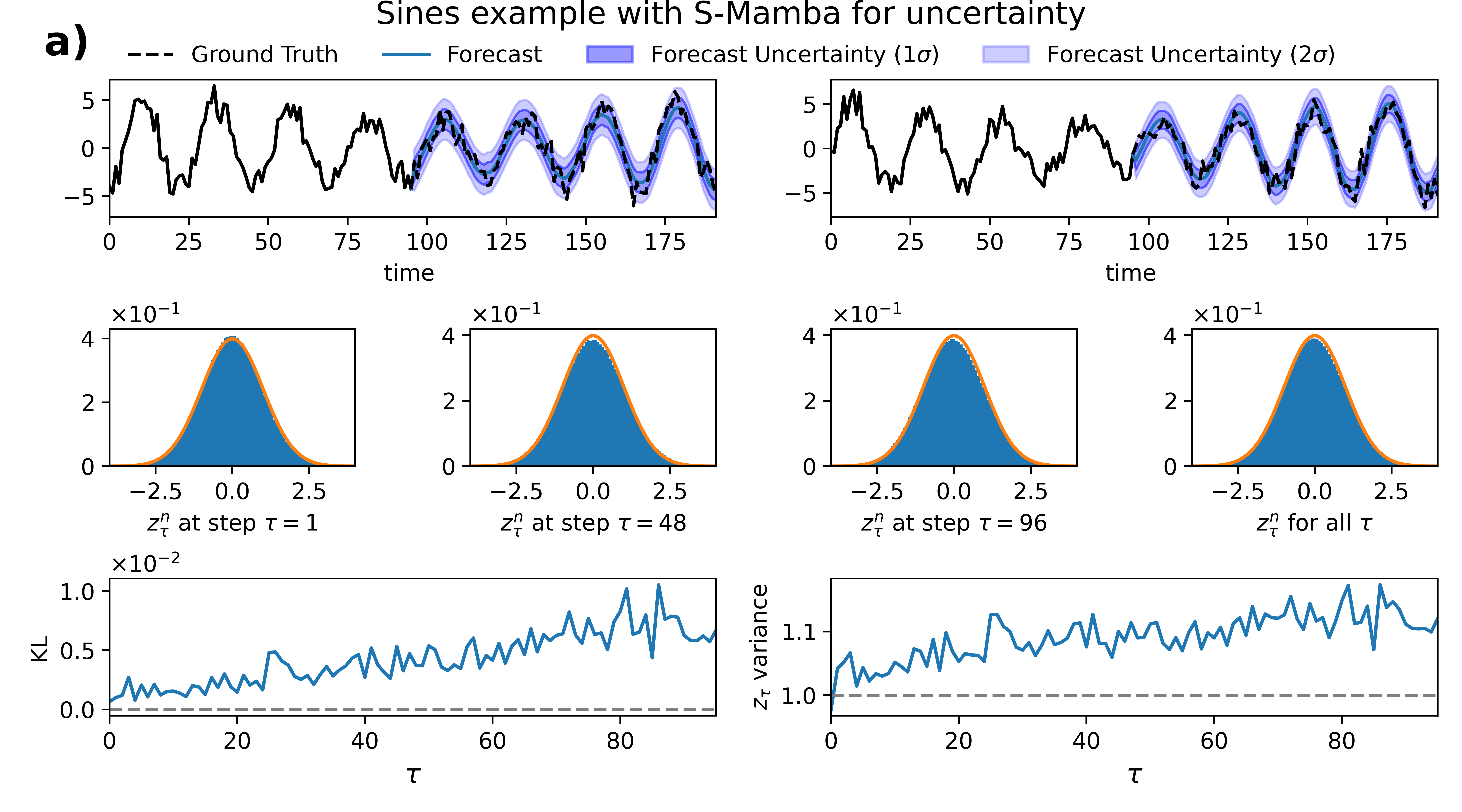}    \includegraphics[width=0.45\linewidth]{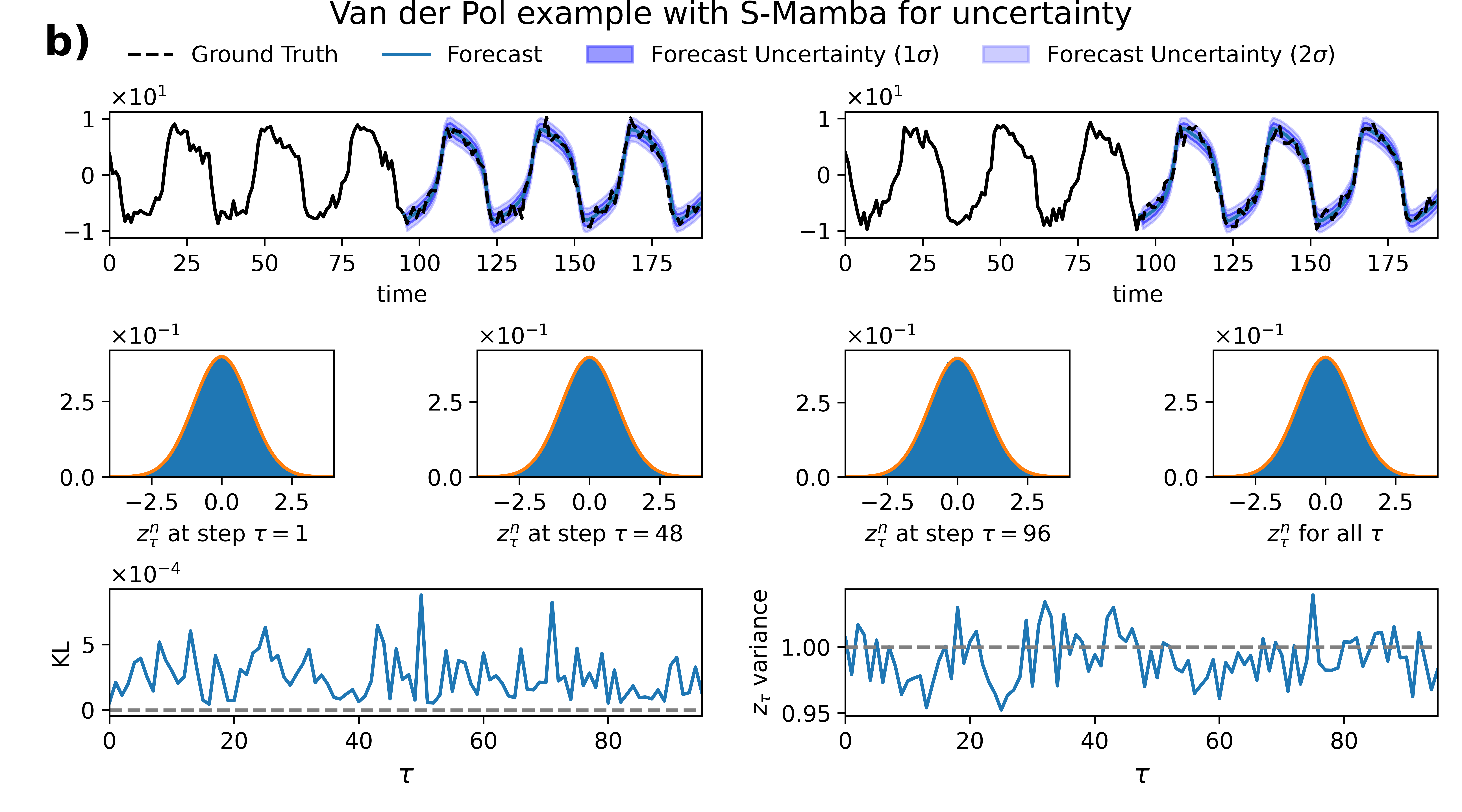}
    \includegraphics[width=0.45\linewidth]{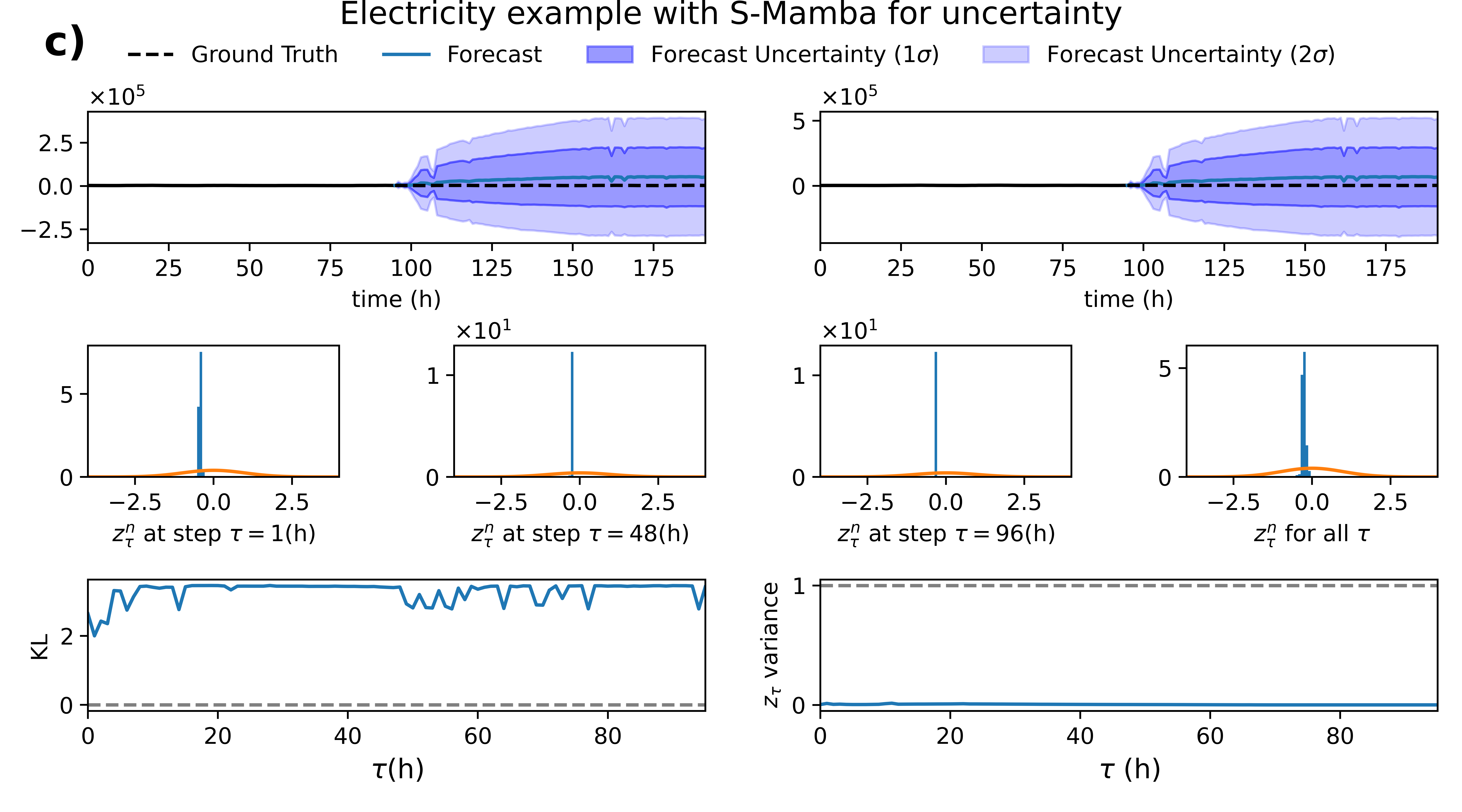}    \includegraphics[width=0.45\linewidth]{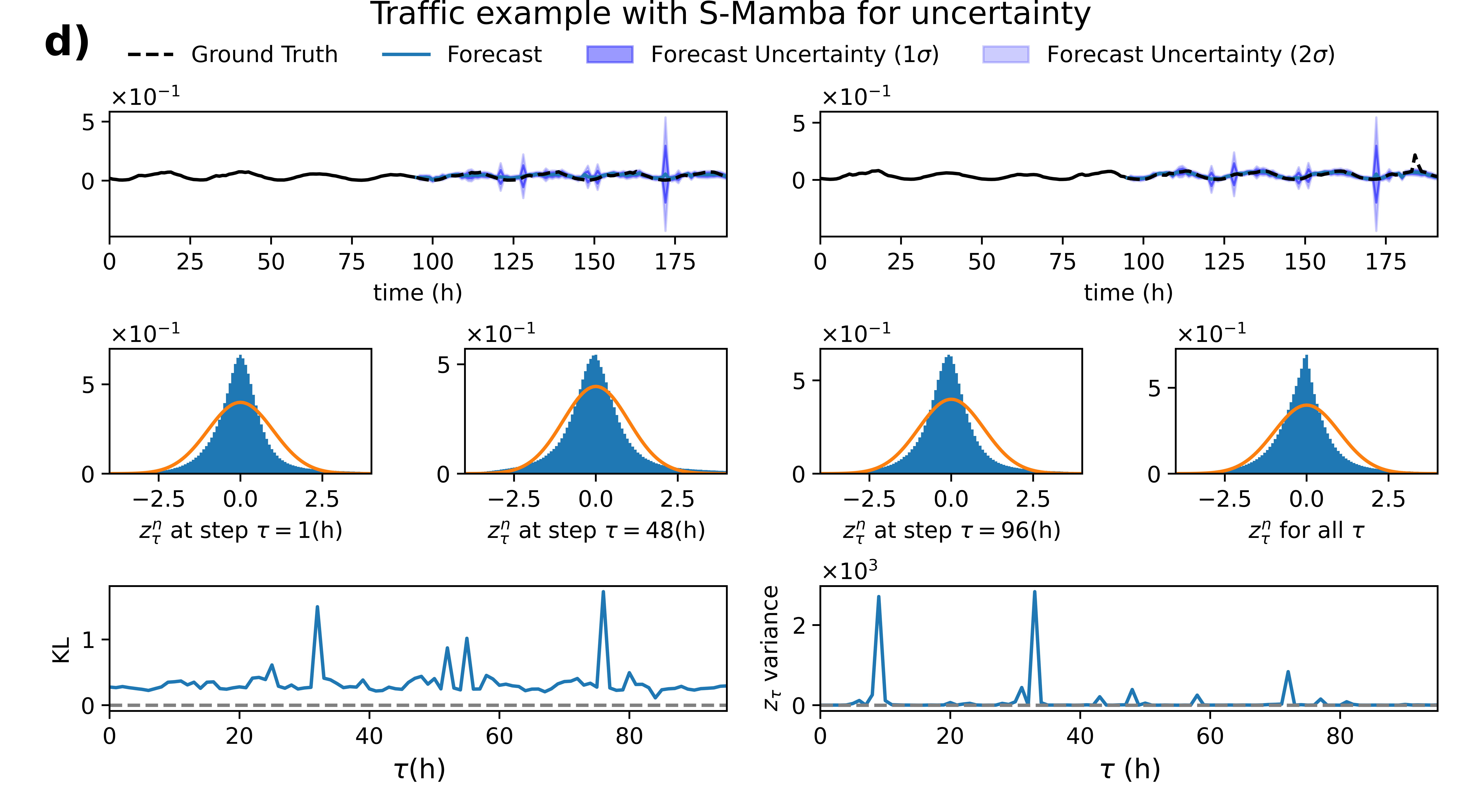}
    \vspace{-.5cm}
    \caption{\textbf{Effect of using S-Mamba with a softplus activation for $\sigma^{\phi_\sigma}_{1:T}(x_{1:P})$ on different datasets.}  
    Unlike the main text results, where a fully connected neural network was used for $\sigma^{\phi_\sigma}_{1:T}(x_{1:P})$, here we employ S-Mamba with a softplus activation.  
    Panels a) and b) correspond to the sines (Fig.\ref{fig:sines}) and Van der Pol (Fig.\ref{fig:VdP}) examples, respectively.  
    In these cases, the KL divergence and variance of standardized residuals $z_\tau^n$ remain of the same order as in the main text.
    However, in panel c), which corresponds to the electricity dataset, S-Mamba dramatically overestimates the uncertainty (compared to Fig. \ref{fig:electricity}), indicating a failure of S-Mamba as an architecture for the uncertainty.  A similar overestimation, although of lesser extend, is observed in the traffic dataset (panel d). }
    \label{fig:unc_smamba}
\end{sidewaysfigure}

 The updated results are presented in Fig. \ref{fig:unc_smamba}, where panels a, b,  c, and d correspond to the sines, Van der Pol, electricity, and traffic datasets, respectively. The only difference from the main text is the substitution of the fully connected network with S-Mamba for $\sigma^{\phi_\sigma}_{1:T}(x_{1:P})$, while the training setup, described in Sec.\ref{sec:training_steps}. As before, a softplus activation ensures positive values for $\sigma^{\phi_\sigma}_{1:T}(x_{1:P})$.

For both synthetic datasets (Fig. \ref{fig:unc_smamba}a and b), the results are comparable to those obtained with the fully connected network. 
The standardized residuals $z_\tau^n$ closely follow a standard normal distribution, with KL divergence and variance remaining in he same order as in the main text. The variance estimates across different forecast horizons $\tau$ stay within a reasonable range, confirming that S-Mamba can effectively model uncertainty in these controlled synthetic datasets.

However, in the electricity dataset (Fig. \ref{fig:unc_smamba}c), the S-Mamba model severely overestimates the uncertainty, producing error bounds that are unrealistically large. This results in a variance of $z_\tau^n$ that is an order of magnitude larger than 1 and a KL divergence two orders of magnitude higher compared to the fully connected case (Fig. \ref{fig:electricity}).
Similarly, in the Traffic dataset  (Fig. \ref{fig:unc_smamba}d), S-Mamba was even more susceptible to the rare deviations, leading to a KL divergence one order of magnitude higher than the one presented in Fig. \ref{fig:traffic}.
These suggest that while S-Mamba provides a flexible architecture for TSF, it may struggle to produce stable uncertainty estimates in real-world datasets. In short, these uncertainties will not evolve similarly to a latent differential equation as described in Sec. \ref{sec:ssm}.

\section{Results for Brownian motion}\label{SIsec:brownian}

\begin{figure}
    \centering
    \includegraphics[width=0.95\linewidth]{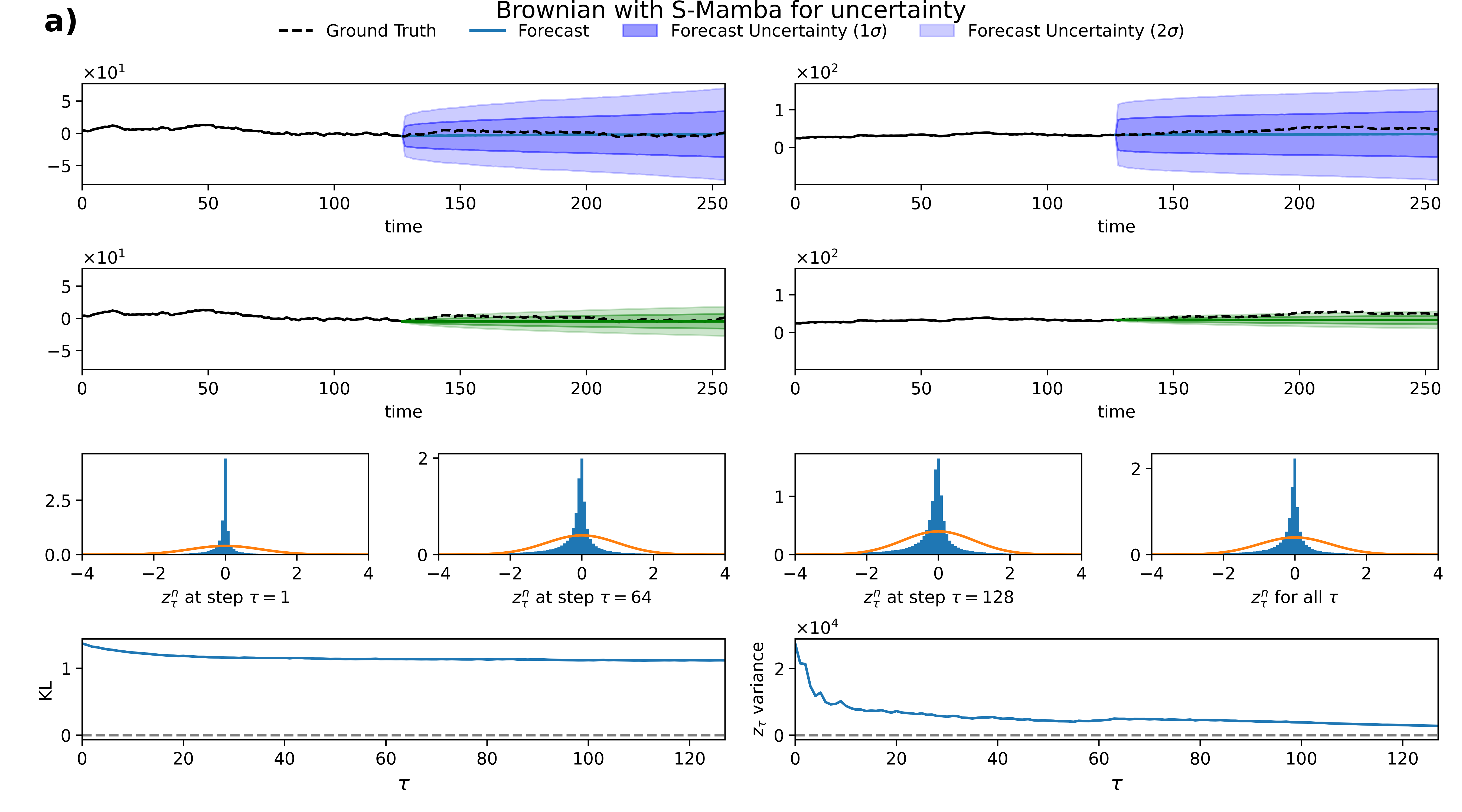}
    \includegraphics[width=0.95\linewidth]{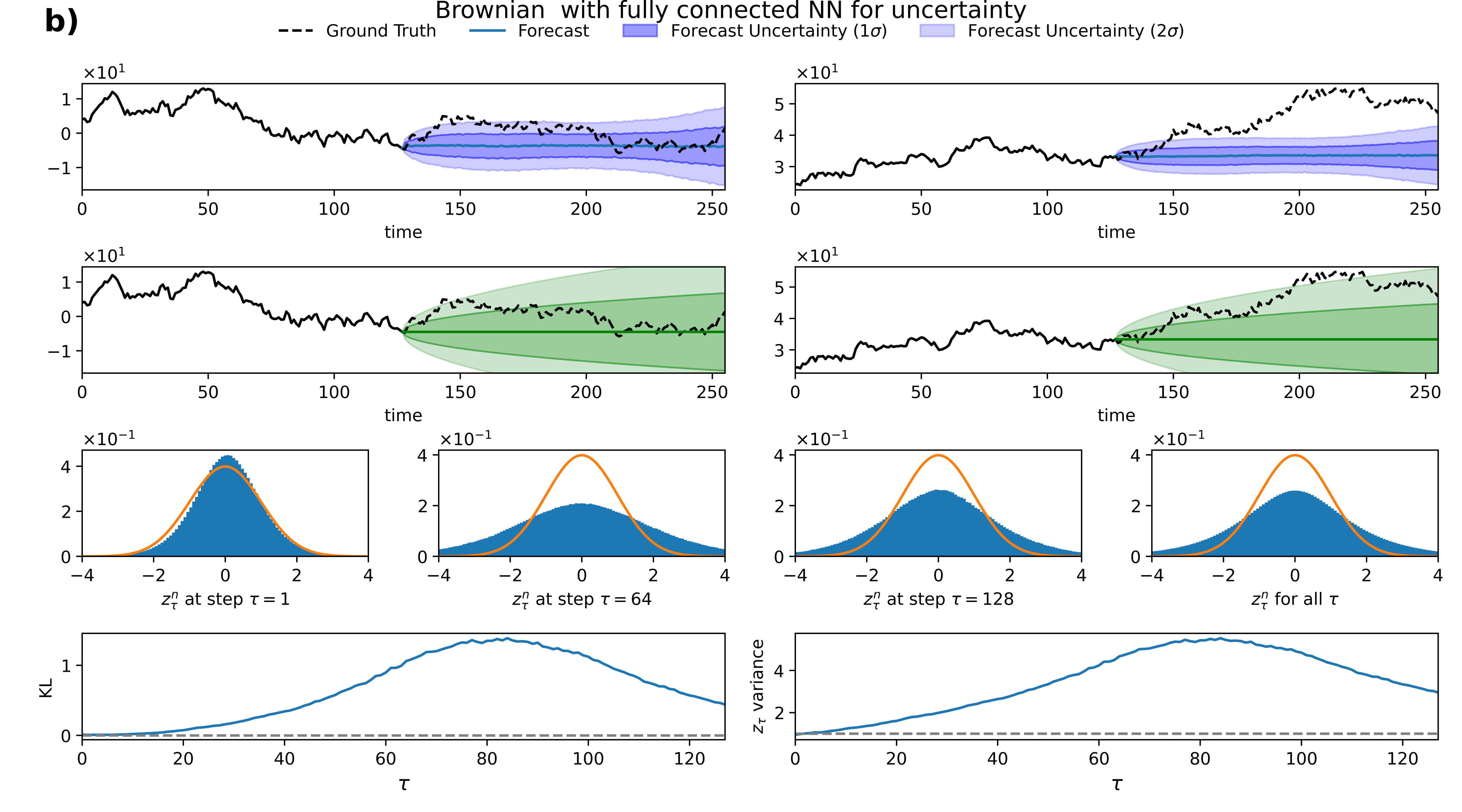}
    \caption{\textbf{Challenges in modeling Brownian motion uncertainty using neural networks.}  
    Brownian motion follows an analytically known probability distribution, where the variance grows as $\sigma^{\phi_\sigma}_{\tau} \sim \sqrt{\tau}$ as expected in \eqref{brownian_real}. The shaded areas represent the predicted distributions (top row) compared to the expected theoretical ones (green shaded on the second row).  
    a) Using S-Mamba for $\sigma^{\phi_\sigma}_{1:T}(x_{1:P})$ results in significantly overestimated uncertainty, producing forecast intervals that are too broad.  
    b) A fully connected network for $\sigma^{\phi_\sigma}_{1:T}(x_{1:P})$ yields more reasonable predictions but still fails to precisely recover the theoretically expected distribution.  
    In both cases, the standardized residuals $z_\tau^n$ show a KL divergence close to 1, and the variance of $z_\tau^n$ reaches up to 5, demonstrating the difficulty of capturing purely stochastic processes with state-space models.}
    \label{fig:brownian}
\end{figure}

As mentioned in Sec. \ref{sec:brownian}, the Brownian motion dataset provides a distinct challenge. Unlike the previous synthetic datasets, where the models successfully learned the relationship between uncertainty and time evolution, here we find that neither S-Mamba nor the fully connected network for $\sigma^{\phi_\sigma}_{1:T}(x_{1:P})$ correctly captures the expected variance growth of $\sigma^{\phi_\sigma}_{\tau}(x_{1:P}) \sim \sqrt{\tau}$ -- see \eqref{brownian_real}. 
As shown in Fig. \ref{fig:brownian}a, when using S-Mamba for variance estimation, the predicted uncertainty is significantly overestimated, leading to an overly broad forecast distribution. 
In contrast, Fig. \ref{fig:brownian}b shows that when using a fully connected network, the model produces more reasonable uncertainty estimates but still fails to match the theoretical distribution. This results in standardized residuals $z_\tau^n$ that deviate from the expected standard normal distribution, with KL divergences close to 1 and variance estimates reaching up to 4.

\pagestyle{empty}

\end{document}